\pdfoutput=1

\documentclass[11pt]{article}

\usepackage[final]{acl}
\usepackage{times}
\usepackage{latexsym}

\usepackage[T1]{fontenc}

\usepackage[utf8]{inputenc}

\usepackage{microtype}

\usepackage{inconsolata}

\usepackage{graphicx}

\usepackage{booktabs}
\usepackage{multirow}
\usepackage{xcolor}
\usepackage{colortbl}
\usepackage{array}
\usepackage{amsmath}
\usepackage{amssymb}
\usepackage{tabularx}
\usepackage{makecell}
\usepackage{listings}
\usepackage{textcomp}
\usepackage{multicol}

\title{TripTailor: A Real-World Benchmark for Personalized Travel Planning}

\author{Yuanzhe Shen\thanks{\ \ These authors contributed equally.}, Kaimin Wang\footnotemark[1], Changze Lv, {\bf Xiaoqing Zheng\thanks{\ \ Corresponding author.}, Xuanjing Huang} \\
  School of Computer Science, Fudan University, Shanghai, China \\
  \texttt{\{yzshen25,kmwang22\}@m.fudan.edu.cn} \\
 \texttt{\{zhengxq,xjhuang\}@fudan.edu.cn} \\}

\begin{document}
\maketitle
\begin{abstract}
The continuous evolution and enhanced reasoning capabilities of large language models (LLMs) have elevated their role in complex tasks, notably in travel planning, where demand for personalized, high-quality itineraries is rising. However, current benchmarks often rely on unrealistic simulated data, failing to reflect the differences between LLM-generated and real-world itineraries. Existing evaluation metrics, which primarily emphasize constraints, fall short of providing a comprehensive assessment of the overall quality of travel plans. To address these limitations, we introduce TripTailor, a benchmark designed specifically for personalized travel planning in real-world scenarios. This dataset features an extensive collection of over 500,000 real-world points of interest (POIs) and nearly 4,000 diverse travel itineraries, complete with detailed information, providing a more authentic evaluation framework. Experiments show that fewer than 10\% of the itineraries generated by the latest state-of-the-art LLMs achieve human-level performance. Moreover, we identify several critical challenges in travel planning, including the feasibility, rationality, and personalized customization of the proposed solutions. We hope that TripTailor will drive the development of travel planning agents capable of understanding and meeting user needs while generating practical itineraries.\footnote{Our code and dataset are available at \url{https://github.com/swxkfm/TripTailor}.}
\end{abstract}

\section{Introduction}
The field of artificial intelligence has seen remarkable advancements in recent years, particularly with the evolution of large language models (LLMs) \citep{achiam2023gpt,yang2024qwen2,liu2024deepseek}. These sophisticated models have enhanced reasoning capabilities and a remarkable ability to generate human-like text, making them invaluable across various applications. One such application is in travel planning, where AI-powered tools are revolutionizing how individuals and organizations organize their trips \citep{Roadtrippers,Layla}. 

While LLMs hold significant potential for travel planning, current systems mainly rely on numerous rule combinations and human intervention. Achieving fully autonomous planning agents that can generate feasible, rational and personalized itineraries remains a considerable challenge. In a benchmark test for domestic travel planning in the U.S. called TravelPlanner \citep{TravelPlanner}, even the most advanced model at the time, GPT-4, achieved a mere 0.6\% success rate when adhering to all constraints. Although this initial finding was disappointing, subsequent research rapidly advanced the field. For instance, \citet{gundawar2024robust} introduced the LLM-Modulo framework for travel planning, which iteratively combines LLMs with a series of external verifiers, increasing the success rate to 20.6\%. Meanwhile, \citet{hao2024large} developed a strategy that integrates LLM-based and algorithm-based planning methods, substantially raising the success rate to 97\%, effectively addressing the challenge. While methods employing formal verification tools significantly enhance LLMs' ability to manage complex constraints, does this mean that travel itineraries produced by such LLMs can compete with those carefully designed by humans in real-world scenarios? 

\begin{figure*}[t]
   \centering
   
\includegraphics[width=\textwidth]{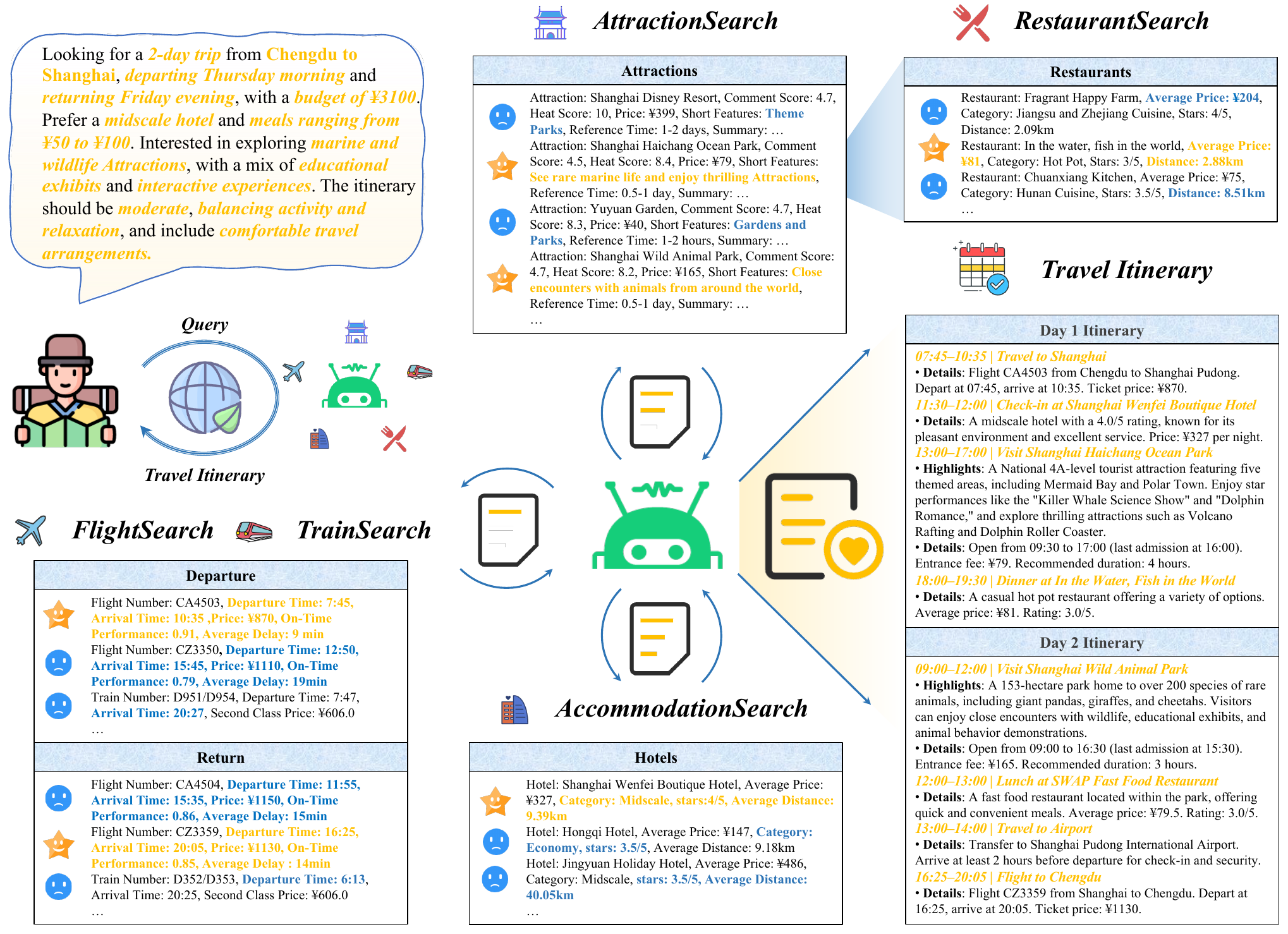}
\vspace{-.3in}
   \caption{Overview of TripTailor. Given a query, language agents utilize a range of tools to gather, filter, and integrate relevant information, thereby gradually formulating a comprehensive travel itinerary. These agents are expected not only to ensure the feasibility and rationality of the itinerary from multiple dimensions, but also to analyze the user's specific needs and personalized preferences in depth, providing a tailor-made travel plan. 
   }
   \label{overview}
   
\vspace{-.15in}
\end{figure*}

To accurately assess this, we need to perform a comparative analysis of different plans related to each user query. However, existing benchmarks for evaluating travel itinerary, such as TravelPlanner \citep{TravelPlanner} and ChinaTravel \citep{shao2024chinatravel}, exhibit certain limitations in terms of authenticity and data scale. TravelPlanner relies primarily on simulated data for its evaluations, making it difficult to reflect actual conditions. In contrast, while ChinaTravel utilizes real data, it only covers 10 cities and about 1,200 POIs per city, which is inadequate to capture the complexities of actual travel needs. Furthermore, current evaluation frameworks overly emphasize specific constraints, making them not only poorly scalable but also incapable of assessing the overall quality of travel plans.

To address the challenges mentioned above, we introduce TripTailor, a novel large-scale travel dataset and an accompanying evaluation framework. This comprehensive dataset includes 40 of China's most popular tourist cities, with an average of 12,500 POIs in each city. It also contains nearly 4,000 detailed samples of real travel itineraries, reflecting a variety of travel scenarios. Our evaluation results indicate that the performance of LLMs in travel planning remains below human standards. Adhering to fundamental constraints ensures only the feasibility and logical accuracy of the plans, but does not guarantee their rationality or efficiency. For example, proposed routes might include unnecessary detours, or the time allocated for each POI may not be optimally planned. In addition, LLMs cannot fully address the diverse and personalized preferences of users. Therefore, although technological advancements have brought LLMs closer to practical applications in travel planning, there remains a gap in fully capturing the nuanced considerations that are inherent in human planners. In summary, our main contributions are as follows:

\noindent\textbf{1. Comprehensive Travel Dataset:} TripTailor is constructed from real-world data sources, boasting a scale that exceeds existing datasets by more than an order of magnitude, thus offering a more diverse testing environment for comprehensive evaluation. Most importantly, it contains over 4000 pairs of real user travel needs and corresponding itineraries, providing valuable insights into traveler preferences and high-quality travel plans.

\noindent\textbf{2. Integrated Evaluation Framework:} We introduce a novel framework to assess the feasibility, rationality, and personalization of travel itineraries through three distinct methodologies: objective metrics, LLM-based evaluation, and a specialized reward model. To the best of our knowledge, this represents the first systematic approach to comparatively evaluate LLM-generated itineraries against real-world travel plans.

\noindent\textbf{3. Workflow Framework for Travel Planning:} We propose a workflow decomposition method that mimics human travel planning processes, serving as a baseline approach. By breaking down the key steps involved in itinerary design, our method facilitates the creation of rational travel plans. 

\section{Related Work}

\subsection{LLM-based Agents for Planning}

As the reasoning capabilities of LLM-based agents continue to improve, these systems have demonstrated unprecedented application potential in various fields such as healthcare \citep{qiu2024llm}, education \citep{zhang2024simulating}, and finance \citep{xing2024designing}. A pivotal aspect of these agents is their planning ability, which involves complex information processing, logical reasoning, decision making, and adaptive adjustment based on the feedback received. To further enhance these capabilities, researchers have developed various strategies, such as task decomposition \citep{shen2023hugginggpt,wang2023plan,singh2023progprompt} and reasoning enhancement \citep{wei2022chain,yao2022react,yao2024tree,besta2024graph}. The integration of external validation modules with self-reflection mechanisms \citep{shinn2023reflexion,pmlr-v235-kambhampati24a} has further refined the accuracy of task execution. Despite notable advancements in handling single-objective tasks using these methodologies, agents still confront numerous challenges in addressing complex real-world problems. This is particularly true for multi-objective optimization scenarios, such as travel planning \citep{TravelPlanner}, which require a comprehensive consideration of multiple interrelated factors.


\subsection{Evaluation of LLMs' Planning Capability}

Evaluating the planning abilities of LLM-based agents is a key topic in current research. Traditionally, studies in this field have focused on domains with clear, easily quantifiable objectives, such as coding and software development \citep{liu2023agentbench,zhang2024codeagent}, web interactions \citep{zhou2023webarena,deng2024mind2web}, tool usage \citep{ruan2023tptu,qin2023toolllm,du2024anytool}, and gaming environments \citep{wang2022scienceworld}. In contrast, travel planning stands apart from these tasks, as it cannot be measured through criteria that include code compilation, goal achievement, or score-based evaluations according to predefined rules. Travel planning is a deeply personalized task, where the effectiveness and quality of plans are highly subjective and vary widely among individuals. Previous studies have evaluated these plans by predefining a set of consensus constraints \citep{TravelPlanner} or manually scoring a limited number of cases \citep{chen2024travelagent}. However, these approaches are limited in terms of scalability and their ability to provide a comprehensive evaluation of overall plan quality and alignment with user needs. To address these issues, we propose a comprehensive evaluation method that utilizes LLMs and incorporates a reward model designed to test whether agents can generate human-level travel plans.

\section{Benchmark}

\begin{table*}[htbp]
   \centering
   \begin{tabular}{lp{10cm}}
      \toprule
      \textbf{Evaluation Metrics} & \textbf{Description} \\
      \midrule
      \multicolumn{2}{c}{\cellcolor{lime!10}\textbf{Feasibility}\vspace{0.25em}} \\ \hline
      Within Sandbox & All information in the plan must be within the closed sandbox; otherwise, it will be considered a hallucination. \\
      Complete Information & No key information should be left out of the plan, such as the lack of accommodation during travel. \\
      \midrule
      \multicolumn{2}{c}{\cellcolor{cyan!10}\textbf{Rationality}\vspace{0.25em}} \\ \hline
      Diverse Restaurants & Restaurant choices should not be repeated throughout the trip. \\
      \textcolor{orange}{Reasonable Meal Prices} & \textcolor{orange}{The selected restaurants should fall within specified price range.} \\
      Diverse Attractions & Attraction choices should not be repeated throughout the trip. \\
      \textcolor{orange}{Appropriate Visit Duration} & \textcolor{orange}{Each attraction should have a visit duration arranged within the recommended time range.} \\
      Defined Budget Limit & The total expenses should not exceed the defined budget. \\
      \textcolor{orange}{Optimized Route} & \textcolor{orange}{The proposed route should minimize travel time between POIs, ensuring an efficient itinerary.} \\
      \midrule
      \multicolumn{2}{c}{\cellcolor{blue!10}\textbf{Personalization}\vspace{0.25em}} \\ \hline
      \textcolor{orange}{Individual Preference} & \textcolor{orange}{The plan should incorporate personalized elements based on the traveler’s interests, preferred cuisine, activities, attractions, desired itinerary intensity, spending habits, and other relevant factors.} \\
      \bottomrule
   \end{tabular}
   \caption{Evaluation metrics. Metrics in black originate from TravelPlanner, whereas metrics in \textcolor{orange}{orange} are our newly proposed measurements designed to evaluate the overall rationality and personalization of travel plans. }
   \label{Metrics}
\vspace{-.15in}
\end{table*}

\subsection{Environment Introduction}
TripTailor provides a comprehensive sandbox environment dedicated to travel planning. This sandbox covers 40 of the most popular tourist cities in China and offers a wide range of travel options. These options encompass 28,832 train schedules and 15,110 flight routes, complete with precise information on departure and arrival times, ticket prices, and on-time performance. The sandbox features 5,622 curated attractions, each with user ratings, popularity indices, ticket prices, geographical locations, key highlights, and recommended visiting durations. In addition, it integrates information on 89,224 hotels and 422,120 restaurants, providing details on categories, user ratings, prices, and locations. For more details on the sandbox and tools, please refer to Appendix~\ref{Additional Benchmark Details}.

\subsection{Benchmark Construction}

\paragraph{Step I: Sandbox Environment Establishment.} We collect information from the open Internet about attractions, hotels, and restaurants in 40 cities across China, along with flight and train schedules between these cities over the course of a week. To guarantee the quality of our dataset, we retain only attractions rated 4A or higher, or those mentioned in verified travel plans, supplementing them with specific characteristics and brief descriptions. In addition, we exclude hotels and restaurants that lack price information, category details, or ratings. Lastly, we utilize Amap, a widely-used mapping service to fill in any missing latitude and longitude coordinates for POIs.

\paragraph{Step II: Realistic Travel Itinerary Construction.} We gather self-guided travel itineraries from online travel agencies, carefully selecting those that offer comprehensive details and have received high ratings to ensure the quality of executable plans. To facilitate the planning of feasible trips, we selected 10 major cities as departure points, while destination cities remain 40. The reason for this design is the lack of direct transportation options between many non-major cities, and considering transfers introduces unnecessary complexity. For each itinerary, a departure city and a departure date are randomly assigned, where the departure date can be any day of the week. The transportation details for the first day of travel and the final day of return are populated using relevant flight or train options from the sandbox. When meals are not included in the itinerary, we find suitable restaurants near the most recently visited POIs to optimize the travel plan. Next, we extract POIs based on time slots designated for each day of the itinerary, supplementing their prices, categories, and descriptions with information from the sandbox. Finally, we utilize an LLM to rewrite the itineraries into coherent, detailed plans with explicit timelines. Note that the LLM's role is strictly limited to reorganizing and refining the existing, verified itinerary information into coherent and reader-friendly narratives. During this process, it may make minimal temporal adjustments to ensure natural progression, but crucially, it does not fill missing information from its pretraining knowledge.




\paragraph{Step III: User Query Construction.} For each itinerary, we extract relevant information, including departure and return dates, durations, and hotel ratings. Additionally, we calculate the cost range for meals and overall budgets. These elements, along with the travel plan, are submitted to an LLM that generates user queries based on the provided details. We also prompt the model to create first-person conversational expressions, focusing on high-level abstract concepts such as types of activities, types of attractions, types of cuisine, and the intensity of travel, rather than specific details about POIs.
The queries are categorized into two distinct levels of difficulty: travel itineraries spanning 2-3 days are classified as \textit{Easy}, whereas those spanning 4 to 7 days are designated as \textit{Hard}. For more information on the dataset distribution and query generation, please refer to Appendix~\ref{Additional Benchmark Details} and ~\ref{sec: query}.

\paragraph{Step IV: Quality Control.} 
We extract key information from established travel plans, evaluate their feasibility and rationality, and identify problematic days with missing or flawed itineraries for regeneration. If the issues persist, the plan is discarded. Subsequently, we utilize an LLM to analyze the coherence between the generated queries and the travel plans. If the score is low and the regenerated queries still do not meet the requirements, the plan is removed. Finally, we manually review the key information in the test set to ensure overall quality, including anomalies at itinerary end (e.g., redundant post-return activities), transportation-activity timing (e.g., insufficient flight buffers), scheduling rationality (e.g., disruptive midnight activities), idle time control (e.g., excessive gaps), and activity selection conflicts (e.g., unresolved overlaps in multi-option slots).

\subsection{Evaluation}
\subsubsection{Plan Quality Assessment}
As illustrated in Table \ref{Metrics}, we evaluate the quality of a plan from three dimensions: feasibility, rationality, and personalization. For the objective standards of feasibility and rationality, we utilize an LLM to extract key elements from the natural language descriptions of travel plans, including the locations and types of activities scheduled for each time range daily. We then match this information within our sandbox environment, and the evaluation process is completed through automated scripts. As for personalization, we assess it directly based on the natural language description of the travel plan. Specifically, our newly proposed evaluation metrics and methods are described as follows:

\paragraph{Optimized Route.} 
When planning a trip, the ideal strategy is to arrange geographically close POIs in a continuous itinerary for the same day, which helps to reduce travel time and improve efficiency. To quantitatively assess the transportation efficiency of the plan, we use the latitude and longitude coordinates of each consecutive POI in the extracted daily itinerary to calculate the average distance for each segment, denoted as \( D_{\text{avg}} \). 
\[
D_{\text{avg}} = \frac{\sum_{k=1}^{n_d} \left( \frac{\sum_{j=1}^{M_k-1} d_{j,j+1}^k}{M_k - 1} \right)}{n_d}
\]

\noindent where \( n_d \) is the total number of days in the itinerary, \( M_k \) is the number of POIs for day \( k \), and
\( d_{j,j+1}^k \) is the distance between consecutive POIs on day \( k \).

\begin{table*}[!ht]
\centering
\resizebox{\textwidth}{!}{ 
\begin{tabular}{l *{8}{c}}
\toprule
 & Average Route & \multicolumn{2}{c}{Feasibility} & \multicolumn{2}{c}{Rationality} & \multicolumn{2}{c}{Personalization} & Final\\
 & Distance Ratio & \multicolumn{2}{c}{Pass Rate} & \multicolumn{2}{c}{Pass Rate} & \multicolumn{2}{c}{Surpassing Rate} & Surpassing Rate\\
\cmidrule(lr){3-4} \cmidrule(lr){5-6} \cmidrule(l){7-8}
  & & Micro & Macro & Micro & Macro & LLM & RM \\
\midrule
 \rowcolor{lightgray}\multicolumn{9}{c}{\textit{\textbf{Easy} (\#354)}}\\
\midrule
Workflow$_{\rm GPT-4o\ mini}$ & 1.8 & 98.9 & 98.6 & 94.3 & 74.3 & 14.1 & 11.6 & 18.1 \\
Direct$_{\rm Qwen2.5-7b-Instruct}$ & 4.0 & 73.9 & 59.0 & 66.4 & 7.3 & 3.9 & 3.4 & 1.1 \\
Direct$_{\rm Qwen2.5-32b-Instruct}$ & 3.6 & 91.8 & 83.9 & 74.1 & 17.8 & 13.8 & 17.8 & 7.1 \\
Direct$_{\rm GPT-4o\ mini}$ & 4.4 & 93.1 & 87.0 & 72.0 & 11.6 & 6.2 & 8.8 & 1.4 \\
CoT$_{\rm GPT-4o\ mini}$ & 4.4 & 92.1 & 85.0 & 75.2 & 16.4 & 7.6 & 8.8 & 3.4 \\
ReAct$_{\rm GPT-4o\ mini}$ & 4.3 & 85.3 & 77.7 & 74.4 & 16.4 & 8.8 & 3.4 & 2.3 \\
Reflexion$_{\rm GPT-4o\ mini}$ & 4.2 & 85.6 & 79.4 & 72.9 & 15.0 & 8.5 & 2.5 & 0.3 \\
Direct$_{\rm DeepSeek-V3}$ & 3.8 & 97.3 & 94.9 & 77.9 & 22.9 & \textbf{30.5} & 17.0 & 11.9 \\
Direct$_{\rm GPT-4o}$ & \textbf{3.4} & \textbf{97.7} & \textbf{95.5} & \textbf{80.2} & 28.8 & 17.8 & \textbf{18.4} & 10.2 \\
Direct$_{\rm o1-mini}$ & 3.6 & 91.0 & 83.9 & 78.7 & \textbf{33.3} & 29.1 & 9.6 & \textbf{16.1} \\

\midrule
 \rowcolor{lightgray}\multicolumn{9}{c}{\textit{\textbf{Hard} (\#349)}}\\
\midrule
Workflow$_{\rm GPT-4o\ mini}$ & 1.7 & 97.7 & 96.0 & 88.8 & 53.0 & 17.5 & 12.0 & 14.3 \\
Direct$_{\rm Qwen2.5-7b-Instruct}$ & 3.5 & 63.8 & 40.4 & 60.7 & 2.3 & 1.4 & 6.0 & 0.3 \\
Direct$_{\rm Qwen2.5-32b-Instruct}$ & 3.3 & 88.0 & 77.1 & 62.3 & 6.3 & 7.4 & 14.9 & 2.3 \\
Direct$_{\rm GPT-4o\ mini}$ & 3.6 & 85.5 & 71.6 & 64.6 & 1.7 & 3.7 & 11.5 & 0.6 \\
CoT$_{\rm GPT-4o\ mini}$ & 3.5 & 85.2 & 72.2 & 64.4 & 1.1 & 3.7 & 8.0 & 0.3 \\
ReAct$_{\rm GPT-4o\ mini}$ & 3.2 & 84.5 & 77.4 & 60.2 & 2.9 & 4.3 & 1.4 & 0.0 \\
Reflexion$_{\rm GPT-4o\ mini}$ & 3.5 & 86.5 & 78.2 & 61.7 & 3.4 & 4.0 & 2.9 & 0.3 \\
Direct$_{\rm DeepSeek-V3}$ & \textbf{3.1} & 95.6 & 91.7 & 69.3 & 7.2 & 14.3 & 24.6 & 3.7 \\
Direct$_{\rm GPT-4o}$ & 3.2 & \textbf{98.9} & \textbf{97.7} & \textbf{73.1} & \textbf{16.3} & 8.0 & \textbf{26.4} & \textbf{4.9} \\
Direct$_{\rm o1-mini}$ & 3.4 & 84.5 & 71.9 & 63.3 & 7.4 & \textbf{17.2} & 15.2 & 2.6 \\

\midrule
 \rowcolor{lightgray}\multicolumn{9}{c}{\textit{\textbf{All} (\#703)}}\\
\midrule
Workflow$_{\rm GPT-4o\ mini}$ & 1.8 & 98.3 & 97.3 & 91.6 & 63.7 & 15.8 & 11.8 & 16.2 \\
Direct$_{\rm Qwen2.5-7b-Instruct}$ & 3.8 & 68.8 & 49.8 & 63.6 & 4.8 & 2.7 & 4.7 & 0.7 \\
Direct$_{\rm Qwen2.5-32b-Instruct}$ & 3.4 & 89.9 & 80.5 & 68.2 & 12.1 & 10.7 & 16.4 & 4.7 \\
Direct$_{\rm GPT-4o\ mini}$ & 4.0 & 89.3 & 79.4 & 68.3 & 6.7 & 5.0 & 10.1 & 1.0\\
CoT$_{\rm GPT-4o\ mini}$ & 4.0 & 88.7 & 78.7 & 69.8 & 8.8 & 5.7 & 8.4 & 1.8\\
ReAct$_{\rm GPT-4o\ mini}$ & 3.8 & 84.9 & 77.5 & 67.3 & 9.7 & 6.5 & 2.4 & 1.1\\
Reflexion$_{\rm GPT-4o\ mini}$ & 3.9 & 86.1 & 78.8 & 67.3 & 9.2 & 6.3 & 2.7 & 0.3\\
Direct$_{\rm DeepSeek-V3}$ & 3.4 & 96.4 & 93.3 & 73.6 & 15.1 & 22.5 & 20.8 & 7.8\\
Direct$_{\rm GPT-4o}$ & \textbf{3.3} & \textbf{98.3} & \textbf{96.6} & \textbf{76.7} & \textbf{22.6} & 12.9 & \textbf{22.3} & 7.5\\
Direct$_{\rm o1-mini}$ & 3.5 & 87.8 & 78.0 & 71.1 & 20.5 & \textbf{23.2} & 12.4 & \textbf{9.4}\\
 
\bottomrule
\end{tabular}
}
\caption{Main results of different LLMs and planning strategies on the TripTailor. Apart from the baseline approach, the best results are marked in bold.}
\label{Main results}
\vspace{-.15in}
\end{table*}

\paragraph{Individual Preference.}
Given the diversity of user needs, along with the inherent complexity of travel planning, a single user query can correspond to multiple high-quality travel itineraries. Under these circumstances, it is unreasonable to expect LLM-generated plans to align precisely with any specific real-world plans, and to compare them directly through a scripted approach. To address this challenge, we have designed two evaluation strategies:

\noindent $\bullet$ \textbf{LLM-as-a-Judge.} For each user query, we provide the actual travel plan as a reference, along with the LLM-generated plan, to an LLM. We then prompt the LLM to compare the two plans across various dimensions, including the selection of hotels, attractions, and restaurants, as well as the depth, breadth, and intensity of the travel experience. After this analysis, the LLM identifies the superior plan and rates each one on a scale of 1 to 5 based on how well it meets user preferences and its overall quality. For more detailed evaluation criteria, please refer to Appendix~\ref{sec:LLM-as-a-Judge}. 

While this approach provides high interpretability, many previous studies have pointed out that LLM-based evaluation systems are often highly sensitive to the positioning of candidate answers \citep{wang2023large,raina2024llm,zheng2023judging}. Moreover, since LLMs can exhibit various inherent biases \citep{liu2023llms,liusie-etal-2024-llm}, depending solely on the outcomes of a single LLM may compromise reliability. To mitigate the potential issues arising from positional bias and inherent biases, we employ two different LLMs for evaluation, alternating the positions of the two options in each assessment, and ultimately taking the average of their results.
    
\noindent $\bullet$ \textbf{Reward-Model-as-a-Judge.} Although this method lacks interpretability, it provides a more nuanced and precise assessment of user preferences. To train the reward model, we first create a pairwise dataset. Specifically, for each user query, the travel itinerary that directly corresponds to that query is designated as a positive sample. Next, we employ the TF-IDF method to retrieve another query from the database that is highly similar to the original query but not an exact match, while ensuring that the departure city, destination city, and travel duration remain the same. The travel itinerary associated with this query is considered a negative sample. The model is then trained using standard methods, with the loss function defined as follows:
\[
    -\mathbb{E}_{(x,y_c,y_r) \sim D} \log\left(\sigma\left(r_\theta(x, y_c) - r_\theta(x, y_r)\right)\right)
\]
where $ D $ is the pairwise dataset, $ r_\theta(x, y) $ is the output of the reward model for user query $ x $ and travel itinerary $ y $ with parameters $ \theta $, $ y_c $ is the preferred travel itinerary, and $ y_r $ is a less preferred one.

\subsubsection{Metrics}

To facilitate a more intuitive comparison and evaluation of the differences between LLM-generated and real plans, we define the following evaluation criteria:

\noindent $\bullet$ \textbf{Feasibility Pass Rate:} This metric assesses the fundamental feasibility of a plan. A plan is deemed infeasible if the LLM cannot produce a valid outcome within 30 steps or if the generated plan includes hallucinations, such as incorrect departure and return details or an inability to match POIs within the sandbox environment.

\noindent $\bullet$ \textbf{Rationality Pass Rate:} Since there is no standard answer for the ``Optimized Route'', we list it separately for reference. The remaining five are utilized to assess the rationality of the plan.

\noindent $\bullet$ \textbf{Personalization Surpassing Rate:} This metric assesses the percentage of LLM-generated plans that surpass real plans in meeting user needs. Results obtained from the LLM and the reward model are presented separately.

\noindent $\bullet$ \textbf{Average Route Distance Ratio:} This metric evaluates the efficiency of a plan. Specifically, we present a ratio of the average distance between consecutive POIs of the LLM-generated plan and the real plan.

\noindent $\bullet$ \textbf{Final Surpassing Rate:} This metric evaluates how well LLM-generated plans match or outperform real plans in terms of personalization, provided that these generated plans satisfy the feasibility and rationality criteria. In particular, an LLM-generated plan is deemed to meet personalization standards if either the score from the LLM or the reward model indicates that it outperforms the real plan.

\section{Experiments}
\subsection{Models and Baselines}

\paragraph{LLMs.} We conducted a comprehensive evaluation of several leading models, both closed-source and open-source, including \href{https://openai.com/api/}{OpenAI GPT-4o}, \href{https://openai.com/api/}{OpenAI GPT-4o mini}, DeepSeek-V3 \cite{liu2024deepseek}, Qwen2.5-7b-Instruct, Qwen2.5-32b-Instruct \cite{yang2024qwen2} and the reasoning model \href{https://openai.com/api/}{OpenAI o1-mini}. 

\paragraph{Methods.}
We test four current mainstream planning methods: Direct, Zero-shot CoT \citep{wei2022chain}, ReAct \citep{yao2022react}, and Reflexion \citep{shinn2023reflexion}. For the same reasons mentioned in TravelPlanner, we do not evaluate tree-based or MCTS methods due to their impracticality for complex tasks like travel planning with large search spaces. We also develop a manual workflow decomposition method to serve as a baseline approach. This method begins by identifying transportation routes between cities, after which the LLM is prompted to rank attractions according to user preferences. Subsequently, the top-ranked attractions are selected, and the LLM generates an initial itinerary. Restaurants near these attractions and the centrally located hotel are then identified based on their geographical proximity and integrated into the plan. Ultimately, a comprehensive itinerary is created based on the detailed information of selected POIs for each day.


\paragraph{Implementation Details.} Given the significantly lower effectiveness of the two-stage mode (tool use and planning) compared to the sole-planning mode observed in TravelPlanner, we provide agents with pre-searched information in all experiments except for the baseline approach to more accurately evaluate agents' planning capabilities rather than their information-gathering capabilities. For LLM evaluation, we employ DeepSeek-V3 and GPT-4o. In the reward model evaluation, we fine-tune Qwen2.5-1.5B-Instruct \citep{yang2024qwen2}. For more implementation details, please refer to Appendix ~\ref{sec:Additional Experiment Details}.

\begin{figure*}[t]
  \centering
  \includegraphics[width=0.9\textwidth]{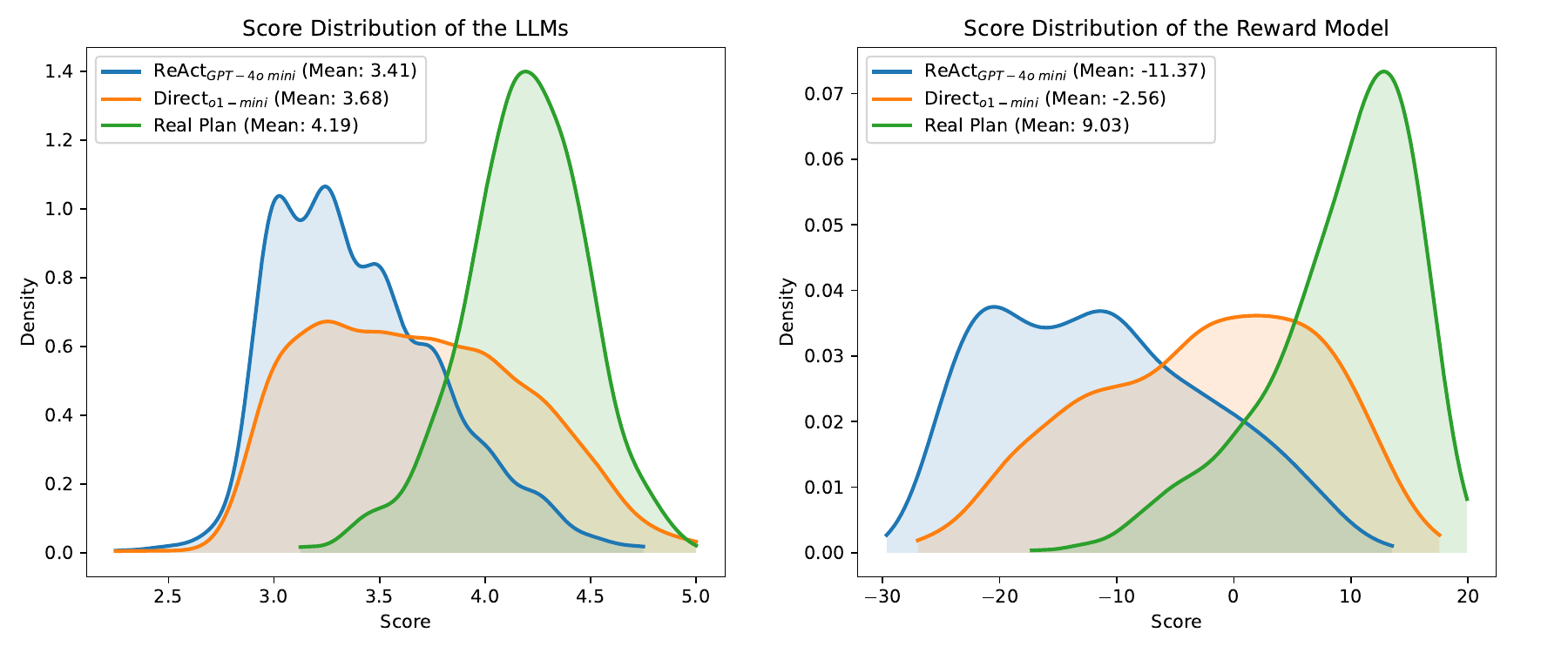}
  \caption{Score distribution in personalized evaluation, shown as KDE plots.}
  \label{score}
\end{figure*}

\subsection{Main Results}

\paragraph{TripTailor provides a challenging benchmark.}
As shown in Table \ref{Main results} and \ref{submetrics}, when provided with complete information on all POIs included in the reference plan, the current state-of-the-art model, GPT-4o, has a success rate of only 21.5\% in generating feasible and rational plans, with a personalization surpassing rate of less than one-third. Notably, even without considering spatial path optimization, only about 7.5\% of the generated plans reach a quality level comparable to real plans. In addition, other existing methods demonstrate more limited performance, failing to generate viable plans that meet the quality of actual plans. These findings highlight the significant challenges faced by current agents in handling complex real-world tasks such as travel planning, particularly in addressing multidimensional constraints and meeting personalized needs.

\begin{table}[t]
\centering
\resizebox{\linewidth}{!}{
\begin{tabular}{lcccc}
\toprule
\multirow{2}{*}{\textbf{Evaluation Metrics}}  & \multicolumn{4}{c}{\textbf{Planning (\textit{All)}}} \\ \cmidrule(l){2-5}
 & \multicolumn{1}{c}{DSV3}  & \multicolumn{1}{c}{GPT-4o}    & \multicolumn{1}{c}{o1-mini} & \multicolumn{1}{c}{Workflow}  \\ 
\midrule
\rowcolor{lightgray}\multicolumn{5}{c}{\textbf{\textit{Feasibility (Pass Rate)}}}    \\ 
\midrule
Within Sandbox  & 93.3 & \textbf{96.6} & 78.0 & 97.3 \\
Complete Information  & 99.6 & \textbf{100.0} & 97.6 & 99.3 \\

\midrule
\rowcolor{lightgray}\multicolumn{5}{c}{\textbf{\textit{Rationality (Pass Rate)}}}    \\ 
\midrule
Diverse Restaurants  & \textbf{98.9} & 98.6 & 73.8 & 99.6\\
Reasonable Meal Prices  & 36.3 & 44.0 & \textbf{48.8} & 98.6\\
Diverse Attractions  & 99.7 & \textbf{99.9} & 91.5 & 96.1\\
Appropriate Visit Duration  & \textbf{65.0} & 64.2 & 61.5 & 72.4\\
Defined Budget Limit  & 68.1 & 76.8 & \textbf{79.8} & 91.3\\

\midrule
\rowcolor{lightgray}\multicolumn{5}{c}{\textbf{\textit{Personalization (Surpassing Rate)}}}    \\ 
\midrule
Individual Preference$_{\rm LLM}$ & 22.5 & 12.9 & \textbf{23.2} & 15.8\\
Individual Preference$_{\rm RM}$ & 20.8 & \textbf{22.3} & 12.4 & 11.8\\
\midrule
\rowcolor{lightgray}\multicolumn{5}{c}{\textbf{\textit{Final}}}    \\ 
\midrule
Final Pass Rate & 14.4 & \textbf{21.5} & 18.3 & 63.3\\
Final Surpassing Rate & 7.8 & 7.5 & \textbf{9.4} & 16.2\\

\bottomrule
\end{tabular}}
\caption{Pass rate and surpassing rate of each evaluation metric. The Final Pass Rate integrates the pass rates of all metrics under Feasibility and Rationality.}
\label{submetrics}
\vspace{-.10in}
\end{table}

\begin{table}[t]
\centering
\resizebox{\linewidth}{!}{
\begin{tabular}{lcccccc}
\toprule
\midrule
\multirow{2}{*}{\textbf{Average}}  & \multicolumn{3}{c}{\textbf{Route Distance (km) }} & \multicolumn{3}{c}{\textbf{Route Distance Ratio }} \\ \cmidrule(l){2-4} \cmidrule(l){5-7}
& \multicolumn{1}{c}{A}   & \multicolumn{1}{c}{A+R}    & \multicolumn{1}{c}{A+R+H} & \multicolumn{1}{c}{A}   & \multicolumn{1}{c}{A+R}    & \multicolumn{1}{c}{A+R+H} \\ 
\midrule
Real Plan & \textbf{5.8} & \textbf{5.5} & \textbf{7.3} & \textbf{1.0} & \textbf{1.0} & \textbf{1.0} \\
Direct$_{\rm DeepSeek-V3}$  & \underline{15.9} & 17.2 & 17.4 & 6.1 & 5.9 & 3.4\\
Direct$_{\rm GPT-4o}$  & 16.1 & \underline{16.7} & \underline{17.1} & \underline{5.6} & \underline{5.5} & \underline{3.3} \\
Direct$_{\rm o1-mini}$ & 17.2 & 17.5 & 18.3 & 6.0 & 5.9 & 3.5  \\
\midrule
\bottomrule
\end{tabular}}
\caption{Comparison of average route distances and distance ratios. A, R, and H denote Attraction, Restaurant, and Hotel, respectively. The smallest results are marked in bold, while underlined values represent the second smallest. Real Plan serves as the baseline.}
\label{distance}
\vspace{-.15in}
\end{table}

\paragraph{Merely satisfying various constraints does not inherently ensure the high quality of a plan or its alignment with user preferences.} As illustrated in Table \ref{submetrics}, When analyzing the plans that fully satisfy all the feasibility and rationality metrics, GPT-4o demonstrates a final pass rate nearly 50\% higher than that of DeepSeek-V3, although their final surpassing rates remain closely aligned. A similar pattern emerges in the performance of the baseline approach: while process decomposition and carefully designed prompts effectively enhance plan feasibility, these improvements do not translate into a notably higher personalization rate. Notably, some plans with slight deviations from constraints (e.g., minor budget overruns) can still be deemed high-quality solutions. This highlights the importance of treating constraints in open-domain tasks, such as travel planning, as flexible rather than rigid requirements. Consequently, merely defining and satisfying easily quantifiable constraints fails to capture the multidimensional nature of human needs and the subtle intricacies of linguistic expression. This underscores the significance of our proposed innovative method for comparative evaluation using real-world plans in TripTailor, which establishes a more challenging and realistic evaluation framework.

\paragraph{Agents struggle to generate feasible, rational and personalized travel itineraries.} In terms of feasibility and rationality, while some agents perform well on micro-level metrics, their scores on macro-level metrics remain relatively low. Moreover, even when the generated plans align with user preferences, they often contain factual inaccuracies or exhibit a lack of rationality. This highlights that current agents frequently make minor errors during the planning process and struggle to comprehensively account for the overall quality of plans.

\subsection{Further Analysis}

\paragraph{Agents struggle to meet personalized needs.} As shown in Figure \ref{score}, real plans achieve superior quality, with 80\% of scores surpassing 4 and an average score of 4.19, positioning their overall quality between ``excellent'' and ``good.'' In contrast, plans generated by o1-mini have an average score of only 3.68, placing them between ``good'' and ``average''. Furthermore, 65\% of these scores fall below 4, and 80\% fail to reach the average level of real plans. Similarly, results from the reward model reveal a consistent trend. These findings suggest that while LLM-generated plans can meet users' basic needs, they still exhibit notable shortcomings in areas such as personalized customization, in-depth experiences, and precise alignment with user preferences.

\paragraph{Agents fail to optimize specific travel itinerary routes.} As shown in Table ~\ref{distance}, LLM-generated plans indicate an average straight-line distance of more than 17 kilometers between POIs, compared to just 7.3 kilometers in real-world plans. The gap further widens to three times when focusing solely on attractions. Such results highlight the inadequacies of current LLMs in spatial-geographic cognition, rendering them incapable of accurately assessing the spatial relationships between POIs during itinerary design. As a result, LLM-generated plans frequently demonstrate two critical shortcomings: first, they often fail to identify geographically proximate attractions for sequential visits, and second, even when adjacent attractions are selected, they are typically not scheduled in temporal proximity. This often leads to significant deviations from optimal routes, resulting in unnecessary increases in transportation time and associated costs.

\paragraph{Reasoning models demonstrate potential in travel planning but still face challenges.} While planning strategies such as ReAct and Reflexion do not show advantages over the Direct method, the reasoning model o1-mini leverages its strong inferential capabilities to excel in itinerary planning tasks, particularly outperforming general LLMs in crafting short-term plans for 2-3 day trips. However, despite o1-mini's leading performance on the test set, its pass rate for the ``Within Sandbox'' criterion is only 78\%, significantly lower than GPT-4o's 96.6\%, highlighting serious hallucination issues. We further identified two primary shortcomings: 1) Fabrication of Information: The LLM occasionally produces fictitious travel details. 2) Information Confusion: This is the most prevalent issue. The LLM sometimes confuses or misapplies transportation arrangements and fails to accurately differentiate between categories such as attractions, hotels, and restaurants. For example, it erroneously lists a restaurant as an attraction in the itinerary.

\section{Conclusion}
We introduce TripTailor, a benchmark specifically designed to evaluate the performance of agents in real-world travel planning scenarios. By collecting extensive real-world POIs and travel itineraries, and adopting a comparative evaluation approach, we effectively address the issues of limited authenticity, incomplete evaluation, and poor scalability present in previous benchmarks. Experiments reveal that even state-of-the-art models struggle to generate feasible, rational, and personalized travel itineraries. We hope that our work provides valuable insights for future research and advances the development of smarter travel planning agents.
\section*{Limitation}
Our work primarily focuses on travel scenarios within China. Due to differences in travel habits and cultural backgrounds around the world, TripTailor may not fully address the specific needs of users in other countries, which somewhat limits its global applicability. However, we believe that the inclusion of 40 cities and nearly 4,000 diverse travel plans can provide a comprehensive overview of the variations in travel plan and the diverse characteristics of travel demand.

Another limitation stems from our query generation process: despite a dual-review mechanism for quality control, the simulated travel plan queries tend to reflect LLM preferences rather than real-world user behavior. Additionally, queries are limited to single-turn interactions, while authentic travel planning often involves iterative, multi-turn dialogues with less detailed initial requests.

In terms of evaluation methods, our personalized and quality assessments primarily depend on LLMs and reward models, which inevitably introduce model bias and hallucinations. To mitigate these issues, we have implemented various corrective measures and validated the effectiveness of our evaluation methods through manual sampling assessments. Designing more objective evaluation metrics and training more robust evaluation models may be important directions for future work.


\bibliography{main}

\appendix
\clearpage

\begin{table*}[t]
    \centering
    \begin{tabularx}{0.9\linewidth}{ccX} 
        \toprule
        \midrule
         POI & Counts & \multicolumn{1}{c}{Information}  \\
         \midrule
         Attractions & 5622 & Poi Id, City, Poi Name, Comment Count, Comment Score, Heat Score, Sight Level Str, Price, Latitude, Longitude, Tag Name List, Short Features, Reference Time, Summary, Opening Hours \\
         \midrule
         Accommodations & 89224 & Name, Real City, Avg Price, Small Cate, Stars, Review Count, Good Remarks, Bad Remarks, Longitude, Latitude, Product Rating, Environment Rating, Service Rating \\
         \midrule
         Restaurants & 422120 & Name, Real City, Avg Price, Small Cate, Stars, Review Count, Good Remarks, Bad Remarks, Longitude, Latitude, Product Rating, Environment Rating, Service Rating, Nearby Attractions \\
         \midrule
         Flights & 15110 & Departure City, Arrival City, Distance(Km), Flight Number, Airline, Aircraft Type, Departure Time, Arrival Time, Departure Airport, Departure Airport Latitude, Departure Airport Longitude, Arrival Airport, Arrival Airport Latitude, Arrival Airport Longitude, On Time Performance, Average Delay Minutes, Monday, Tuesday, Wednesday, Thursday, Friday, Saturday, Sunday, Price \\
         \midrule
         Trains & 28832 & Train Number, Station Number, Station Name, Arrival Time, Departure Time, Stop Time, Running Time, Second Class Price, First Class Price, Longitude, Latitude \\
         \midrule
         \bottomrule
    \end{tabularx}
    \caption{POI Information}
    \label{poi_info}
\end{table*}

\begin{table*}[t]
    \centering
    \begin{tabularx}{0.9\linewidth}{lX} 
        \toprule
        \midrule
         Tool &   \multicolumn{1}{c}{Description}  \\
         \midrule
         AttractionSearch & Search for attractions in a given city.\\
         \midrule
         AccommodationSearch & Search for accommodation options near a given coordinate in a given city with a given rating level.\\
         \midrule
         RestaurantSearch & Search for restaurants near a given coordinate in a given city within a given price range.\\
         \midrule
         FlightSearch & Search for available flights between two cities.\\
         \midrule
         TrainSearch & Search for available train routes between two cities.\\
         \midrule
         \bottomrule
    \end{tabularx}
    \caption{Tool Information}
    \label{tool_info}
\end{table*}

\clearpage
\begin{table*}[t]
    \centering
    \begin{tabular}{lcccccc} 
        \toprule
        \midrule
           & 2-day & 3-day & 4-day & 5-day & 6-day & 7-day \\
        \midrule
         Training (\#3,145) & 686 & 1011 & 776 & 498 & 148 & 26 \\
        \midrule
         Test (\#703) & 120 & 234 & 196 & 116 & 29 & 8 \\
        \midrule
        \bottomrule
    \end{tabular}
    \caption{Dataset Distribution}
    \label{dataset_info}
\end{table*}

\begin{table*}[t]
    \centering
    \begin{tabularx}{0.9\linewidth}{lX} 
        \toprule
        \midrule
         Score &   \multicolumn{1}{c}{Description}  \\
         \midrule
         5 (Excellent) & The itinerary exceeds expectations, perfectly aligning with all user preferences. It offers unique, tailored experiences and exceptional value, ensuring a memorable and personalized journey. \\
         \midrule
         4 (Good) & The itinerary largely meets the user’s needs, showing a strong level of personalization and value. However, there may be minor gaps in specific preferences or opportunities for deeper engagement that could enhance the overall experience. \\
         \midrule
         3 (Average) & The itinerary partially satisfies the user’s query, incorporating some preferences but missing key elements in important areas. It fulfills basic requirements but lacks depth, creativity, or engagement in activities, cultural insights, or personalization, resulting in a feeling of generality and mediocrity. \\
         \midrule
         2 (Poor) & The itinerary barely meets expectations, with significant gaps in personalization and relevance. Most elements do not align well with the user’s stated preferences, leading to a less enjoyable and uninspired experience. \\
         \midrule
         1 (Very Poor) & The itinerary fails to address the user's query entirely, displaying no relevance to stated preferences. It is completely generic, offering little to no value or consideration for the user's unique needs and interests. \\
         \midrule
         \bottomrule
    \end{tabularx}
    \caption{Scoring Standard}
    \label{scoring_info}
\end{table*}

\clearpage

\section{Additional Benchmark Details}
\label{Additional Benchmark Details}
For POI information in the sandbox, please refer to Table ~\ref{poi_info}. For tool information, please refer to Table ~\ref{tool_info}. For the data distribution in the training and test sets, please refer to Table ~\ref{dataset_info}.

\section{Additional Experiment Details}
\label{sec:Additional Experiment Details}

\subsection{Pre-searched Information}
All Points of Interest (POIs) specified in real plans are comprehensively incorporated into the provided information, thereby guaranteeing the existence of a feasible solution under all circumstances.

\paragraph{Transportation:} The data is filtered according to the specified departure and destination cities, along with the corresponding departure and return dates. All relevant information is input into LLMs.

\paragraph{Attractions:} For itineraries spanning 2 to 5 days, a random selection of 50 attractions is made. This number increases to 60 and 70 attractions for 6-day and 7-day itineraries, respectively, to accommodate the extended duration.

\paragraph{Restaurants:} A random selection of restaurants is made from the dataset, with the quantity determined by multiplying the number of days by 8. It is ensured that at least 50\% of the selected restaurants fall within the predefined price range. Each restaurant is annotated with information about the five nearest attractions.

\paragraph{Hotels:} From the dataset, 10 hotels are randomly chosen, with the condition that a minimum of 50\% meet both the rate and price range criteria.

\noindent\textbf{Examples of the given information:}
\lstset{
    backgroundcolor=\color[RGB]{250,250,250},
    breaklines=true,
    breakindent=0pt,
    basicstyle=\ttfamily\scriptsize,
    frame=single,
    frameround = tttt,
}\begin{lstlisting}
Attraction Name: Youyang Taohuayuan, 
Level: 5A, Rating: 4.4, Heat Score: 6.2, Price: 63.0, 
Tags: Close to nature; Hidden gems for walking kids; Caves, 
Features: The Peach Blossom Spring in Tao Yuanming's Writings, Recommended Duration: 3-4 hours, 
Opening Hours: Open at 08:00-17:00, 
Summary: 
- Youyang Taohuayuan Scenic Area is a national forest park, a national 5A-level scenic area, and a national outdoor sports training base. It is located in the heart of the Wuling Mountains.
- It integrates karst geological wonders, the agricultural culture of the Qin and Jin dynasties, Tujia ethnic customs, and natural ecological culture, encapsulating the most beautiful primitive scenery.
- The main attractions include Taohuayuan, Fuxi Cave, the ancient city of Youzhou, Taigu Cave, and Taohuayuan Square. The peach orchards in the area are lush and tranquil.
- Among them, Fuxi Cave in the scenic area is about 3,000 meters long, with winding corridors, a deep underground river, and colorful stalactites, presenting a stunning landscape.

Restaurant Name: Xilai Thin Meat, 
Avg Price: 130.5, Category: Korean Cuisine, Rating: 4.5, Good Remarks: 221.0, Bad Remarks: 8.0, 
Product Rating: 8.8, Environment Rating: 8.9, Service Rating: 9.1, 
Nearby Attractions: Guanyinqiao Pedestrian Street; Zhongfu Beicang Cultural and Creative Park; Jiujie Street; Zhou Mansion; Gui Garden

Hotel Name: Chongqing Color Art Hotel, 
Avg Price: 286.5, Category: Upscale, Rating: 4.5, Good Remarks: 33.0, Bad Remarks: 1.0, 
Product Rating: 8.9, Environment Rating: 9.0, Service Rating: 9.0

Flight Number: 3U3003, 
Price: 1110, Departure Time: 15:20, Estimated Arrival Time: 18:30, 
On-Time Performance: 0.94, Average Delay (minutes): 6

Train Number: G309, 
Price: 864, Departure Time: 8:18, Arrival Time: 20:21
\end{lstlisting}

\subsection{Evaluation}
We randomly selected 100 plans from DeepSeek-V3, OpenAI GPT-4o, and OpenAI 01-mini for manual ranking to evaluate performance. The LLM scoring method achieved a precision of 72.22\%, a recall of 61.90\%, and an F1 score of 66.62\%. The RM scoring method recorded a precision of 57.89\%, a recall of 52.38\%, and an F1 score of 54.95\%. The combined LLM + RM method outperformed both, with a precision of 61.29\%, a recall of 90.48\%, and an F1 score of 72.92\%. These results highlight the strong discriminative power of all methods, with the combined LLM + RM approach excelling due to its high recall and F1 score. Thus, we adopted the combined LLM + RM method for comprehensive evaluation, balancing high recall with robust overall performance.

We also recognize the importance of temporal constraints. However, analysis indicates minimal scheduling conflicts (0.57\% average overlap), with negligible impact on overall evaluation. Thus, temporal constraints were not included in the rationality criteria.


\subsubsection{LLM-as-a-Judge}
\label{sec:LLM-as-a-Judge}
For the scoring criteria, please refer to Table ~\ref{scoring_info}. For the specific scoring prompts, please refer to ~\ref{LLM-as-a-Judge Prompt}

\subsubsection{Reward-Model-as-a-Judge}
We fine-tune the Qwen2.5-1.5B-Instruct on 4 RTX 3090 GPUs with the following parameters: batch size of 4, maximum sequence length of 4096 tokens, learning rate of 1e-5, weight decay of 0.01, 2 training epochs, and 2 gradient accumulation steps.

\subsection{Prompts}

\clearpage
\subsubsection{LLM-as-a-Judge Prompt}
\label{LLM-as-a-Judge Prompt}
\lstset{
    backgroundcolor=\color[RGB]{250,250,250},
    breaklines=true,
    breakindent=0pt,
    basicstyle=\ttfamily\scriptsize,
    frame=single,
    frameround = tttt,
    linewidth=\textwidth, 
}\begin{lstlisting}
You are an AI assistant evaluating two travel plans based on following criteria:

Evaluation Criteria and Key Factors to Consider::
- Experiences: Consider both variety and depth. While a diverse range of activities is beneficial, immersive and well-planned experiences that align closely with traveler interests should also be recognized.
- Itinerary Intensity: Evaluate how well the plan matches the traveler's desired itinerary intensity (e.g., relaxed, moderate, packed). Consider the balance between activities and free time, as well as the pacing of the trip.
- Cuisine: Assess the suitability of dining choices to the traveler's stated preferences, including cuisine category and alignment with budget and meal price range.
- Accommodations: Evaluate the quality, comfort, and overall fit with the traveler's stated preferences, including accommodation category and budget range.
- Transportation: Assess the practicality of transportation options with a focus on departure and return times, convenience, cost, and suitability for the traveler's preferences.
- Total Budget Consideration: Staying within the budget is essential, but an itinerary that justifies slightly higher costs through premium experiences is viewed positively, whereas strict cost-cutting at the expense of premium experiences is seen as unfavorable.

Scoring Scale (Out of 5)
1. 5 (Excellent): The itinerary exceeds expectations, perfectly aligning with all user preferences. It offers unique, tailored experiences and exceptional value, ensuring a memorable and personalized journey.
2. 4 (Good): The itinerary largely meets the user's needs, showing a strong level of personalization and value. However, there may be minor gaps in specific preferences or opportunities for deeper engagement that could enhance the overall experience.
3. 3 (Average): The itinerary partially satisfies the user's query, incorporating some preferences but missing key elements in important areas. It fulfills basic requirements but lacks depth, creativity, or engagement in activities, cultural insights, or personalization, resulting in a feeling of generality and mediocrity.
4. 2 (Poor): The itinerary barely meets expectations, with significant gaps in personalization and relevance. Most elements do not align well with the user's stated preferences, leading to a less enjoyable and uninspired experience.
5. 1 (Very Poor): The itinerary fails to address the user's query entirely, displaying no relevance to stated preferences. It is completely generic, offering little to no value or consideration for the user's unique needs and interests.

Output format:
Analysis:
- Personalization Evaluation Analysis: Please analyze each plan first and then provide a rating in JSON format. Based on the Evaluation Criteria and Key Factors to Consider, provide a detailed comparative analysis of how well each plan meets the traveler's preferences and the overall quality of each plan, explaining their strengths and weaknesses.
```json
{
  "Personalization Evaluation": {
    "Scores": {
      "Plan A": X,
      "Plan B": Y,
    }
  }
}
```

Input
- Query: {query}
- Plan A: {plan_a}
- Plan B: {plan_b}
\end{lstlisting}

\clearpage
\subsubsection{User Query Construction Prompt}
\label{sec: query}
\lstset{
    backgroundcolor=\color[RGB]{250,250,250},
    breaklines=true,
    breakindent=0pt,
    basicstyle=\ttfamily\scriptsize,
    frame=single,
    frameround = tttt,
    linewidth=\textwidth,
}\begin{lstlisting}
You are a travel planning assistant skilled in refining travel-related queries to match user itineraries and preferences. Generate query using the input itinerary and preferences provided below.

Key Guidelines  
1. First-Person Query: The query must be phrased as if the user is directly asking for travel recommendations. Do not describe the itinerary. Example: *"I am looking for a 3-day trip..."*, not *"For your 3-day trip, you will..."*.  
2. Align Preferences: Emphasize user interests extracted directly from the itinerary without naming specific attractions, restaurants, or using phrases like *"like [specific name]"*. Focus on types of activities, attraction types, and cuisine types.  
3. Strict Formatting: Output must be a single, concise paragraph phrased as a request without explanations, summaries, or additional formatting.  
4. Budget and Time: Clearly reflect the trip's duration, budget, and meal cost range in simple terms.  
5. Flexibility: Avoid over-specifying times or places to ensure adaptability in planning.  
6. Itinerary Intensity: Based on the input itinerary, determine if the schedule should be relaxed, moderate, or packed and reflect this in the query.  
7. No Specific Place Names: Do not include specific hotels, restaurants, attractions, or districts.  

Mandatory Elements  
- Departure and Return Days: Must exactly match the input Departure and Return Days provided. Do not modify or infer these dates.  
- Departure Time and Return Time: Based on the itinerary, determine if the departure and return times fall into one of the following time ranges:  
  - Early morning (4:00 - 9:00)
  - Late morning (9:00 - 12:00)
  - Afternoon (12:00 - 18:00)
  - Evening (18:00 - 24:00)  
- Trip Duration: Must exactly match the input Trip Duration provided. Do not modify the input duration.
- Departure and Destination Cities: Must explicitly mention both departure and destination cities. These cannot be omitted.
- Hotel Cost Category: Luxury, Upscale, Midscale, or Economy.  
- Budget: Must exactly match the input Budget provided. Do not modify the input budget.  
- Meal Cost Range: Must exactly match the input Meal Cost Range provided. Do not modify or deviate from this range.  
- Desired Itinerary Intensity: Based on the input itinerary, determine if the schedule should be relaxed, moderate, or packed and reflect this in the query. 

Output Instructions  
- Generate only the query as a single, natural-sounding paragraph in first-person.
- Do not include any headings, labels, or additional formatting.  
- Ensure the output is directly usable as a request in itinerary planning systems.  

Example for formatting only  
(The following example demonstrates the correct structure. Do not copy specific names, numbers, or details.)  
```
I am looking for a 4-day trip from Beijing to Tianjin, departing on Monday afternoon and returning on Thursday afternoon, with a budget of 5700. I prefer staying in luxury hotels and dining at restaurants with meal costs over 200. I'm interested in exploring cultural landmarks, historical sites, scenic river cruises, and architectural marvels, along with enjoying diverse cuisines like seafood, Japanese, and Chinese dishes. The itinerary should be moderate in intensity, balancing guided exploration with some downtime."
```

Input   
- Itinerary:  
{itinerary}  
- Trip Duration
{duration}
- Budget:  
{budget}  
- Meal Cost Range:  
{meal_cost_range}  
- Departure and Return Days:  
{departure_and_return_days}

\end{lstlisting}

\clearpage
\subsubsection{Direct Planning Prompt}
\lstset{
    backgroundcolor=\color[RGB]{250,250,250},
    breaklines=true,
    breakindent=0pt,
    basicstyle=\ttfamily\scriptsize,
    frame=single,
    frameround = tttt,
    linewidth=\textwidth,
}\begin{lstlisting}
You are a travel planner tasked with creating a concise and detailed travel plan based on the query and the provided information. The output should include the following sections and adhere to these specifications:  
1. Daily Itinerary  
   - Divide the trip by days.  
   - Specify the current city (e.g., "from A to B" if traveling between cities).  
   - Include exact timings for each activity (e.g., 10:00-12:00).  
   - Ensure sufficient time for travel, meals, rest, and prioritize key activities logically.  
   - Important: All names, prices, and details must strictly match the given information. Do not modify or invent any information.
2. Transportation  
   - Provide flight or train numbers, departure/arrival times, ticket prices, and durations for intercity travel.  
   - First Day Transportation: Specify only the transportation options available on the first day for travel from the departure city to the destination city.
   - Last Day Transportation: Specify only the transportation options available on the last day for travel from the destination city back to the departure city.
3. Accommodation  
   - Specify the hotel name, rating, check-in/check-out times, and average price per night.  
4. Attractions  
   - Detail attraction names, opening hours, entrance fees, recommended duration, and cultural or historical significance.  
5. Dining  
   - List restaurant names, cuisine types, must-try dishes, average cost per person, and operating hours.  
6. Budget Breakdown  
   - Align with the specified budget and provide a cost breakdown for transportation, accommodations, meals, and attractions.  
   - Offer alternatives if the total cost exceeds the budget.  
7. Additional Notes  
   - One Activity per Time Slot: Ensure only one attraction or activity is scheduled in each time slot.  
   - Time Management: Ensure the itinerary is realistic and accounts for travel time between locations. 
   - Distance Consideration: Prioritize activities that are geographically close to each other to minimize unnecessary travel time.
   
Output Formatting  
- Use clear headings (e.g., Day 1 Itinerary).  
- Present information in bullet points or short paragraphs for readability.  
- Ensure alignment with the user's preferences and query context.  
Ensure that the plan is logical, concise, and detailed, while maintaining alignment with the user's budget, interests, and time constraints. The output should avoid unnecessary elaboration or unrelated details.  

Example
Example Query: 
Planning a 3-day trip from Chongqing to Shenyang, departing on Wednesday and returning on Friday, with a focus on exploring historical landmarks, local cuisine, and leisurely shopping in vibrant commercial areas. The trip includes Economy accommodations with breakfast, averaging 74 per night, and a daily budget under 500, with meal costs around 50-70 per person. The itinerary aims to blend cultural experiences, such as visiting museums and ancient architecture, with free time to relax and enjoy the city.

Example Travel Plan:
Day 1 Itinerary: Chongqing to Shenyang 
06:10-11:00 | Travel to Shenyang
Start your journey with a flight on CA4163 from Chongqing to Shenyang. Depart at 06:10 and arrive at 11:00, ensuring a punctual and comfortable trip.
- Ticket Price: 900
---
12:00-12:30 | Check-in at Jijin E-Family Theme Hotel
After arriving in Shenyang, check in at Jijin E-Family Theme Hotel, an Economy hotel with a 3.5 rating. Enjoy a comfortable stay at an average price of 74 per night. Guests have praised the hotel for its pleasant environment and good service.
- Average Price Per Night: 74
---
14:00-16:00 | Explore Taiyuan Street
Spend some time exploring Taiyuan Street, one of Shenyang's most bustling commercial districts. Modeled after Tokyo's Ginza shopping area, it is known as "Northeast China's First Golden Street." The street features a mix of historic Chinese buildings from the 1920s and modern skyscrapers, offering a unique blend of the old and new.
- Opening Hours: All day (Monday-Sunday, January 1-December 31)
- Entrance Fee: Free
- Recommended Duration: 1-3 hours
---
18:00-19:30 | Dinner at Laotieling Shengchuan (Taiyuan South Street Store)
Enjoy a delicious dinner at Laotieling Shengchuan (Taiyuan South Street Store). This popular local chain is known for its tasty skewers, chicken wings, and grilled dishes. A great place to experience local flavors.
- Location: Between Nanba Road and Nanqi Road (next to Xiaotudou)
- Operating Hours: Monday to Sunday, 16:00-02:00
- Average Cost: 70 per person

Day 2 Itinerary
...Itinerary for the Last Day of the Trip...

Example Ends

Given information:{text}
Query: {query}
Travel Plan:
\end{lstlisting}

\clearpage
\subsubsection{ReAct \& Reflexion Planning Prompt}
\lstset{
    backgroundcolor=\color[RGB]{250,250,250},
    breaklines=true,
    breakindent=0pt,
    basicstyle=\ttfamily\scriptsize,
    frame=single,
    frameround = tttt,
    linewidth=\textwidth,
}\begin{lstlisting}
You are a proficient planner. Based on the provided information and query, please give me a detailed plan, including specifics such as flight/train numbers (e.g., F0123456) and cost, restaurant names and cost, hotel names and cost, and attractions names and cost. Note that all the information in your plan should be derived from the provided data. You must adhere to the format given in the example. Additionally, all details should align with common sense. Attraction visits and meals are expected to be diverse. The symbol '-' indicates that information is unnecessary. For example, in the provided sample, you do not need to plan after returning to the departure city. When you travel to two cities in one day, you should note it in the 'Current City' section as in the example (i.e., from A to B). Do not use any Markdown formatting (e.g., do not use `` for bold text). Solve this task by alternating between Thought, Action, and Observation steps. The 'Thought' phase involves reasoning about the current situation. The 'Action' phase can be of two types:
(1) CostEnquiry[Sub Plan]: This function calculates the cost of a detailed sub plan(except transportation cost), which you need to input the people number and plan in JSON format. The sub plan should encompass a complete one-day plan. An example will be provided for reference.
(2) Finish[Final Plan]: Use this function to indicate the completion of the task. You must submit a final, complete plan as an argument.
Example
Query: Could you create a travel plan from Ithaca to Charlotte spanning 3 days, from Wednesday to Friday, with a daily budget under 500 and meal cost range of 50 to 100?
You can call CostEnquiry like CostEnquiry[{{"day": 1,"current_city": "from Ithaca to Charlotte","transportation": "Flight Number: F3633413, from Ithaca to Charlotte, Cost: 450","attraction": "The Charlotte Museum of History, Cost: 10","lunch": "Cafe Maple Street, Cost: 10","dinner": "Bombay Vada Pav, Cost: 15","accommodation": "Affordable Spacious Refurbished Room in Bushwick!, Cost: 250"}}]
You can call Finish like Finish[Day: 1
Current City: from Ithaca to Charlotte
Transportation: Flight Number: F3633413, from Ithaca to Charlotte, Cost: 450
Attraction: The Charlotte Museum of History, Cost: 10
Lunch: Cafe Maple Street, Cost: 60
Dinner: Bombay Vada Pav, Cost: 55
Accommodation: Affordable Spacious Refurbished Room in Bushwick!, Cost: 250

Day 2:
Current City: Charlotte
Transportation: -
Attraction: The Mint Museum, Cost: 10;Romare Bearden Park, Cost: 0
Lunch: Birbal Ji Dhaba, Cost: 66
Dinner: Pind Balluchi, Cost: 67
Accommodation: Affordable Spacious Refurbished Room in Bushwick!, Cost: 250

Day 3:
Current City: from Charlotte to Ithaca
Transportation: Flight Number: F3786167, from Charlotte to Ithaca, Cost: 500
Attraction: Books Monument, Cost: 0
Lunch: Olive Tree Cafe, Cost: 80
Dinner: Kylin Skybar, Cost: 90
Accommodation: -]
Example Ends

{reflections}

You must use Finish to indict you have finished the task. And each action only calls one function once.
Given information: {text}
Query: {query}{scratchpad} 
\end{lstlisting}

\clearpage
\subsubsection{Workflow Planning Prompt}
\lstset{
    backgroundcolor=\color[RGB]{250,250,250},
    breaklines=true,
    breakindent=0pt,
    basicstyle=\ttfamily\scriptsize,
    frame=single,
    frameround = tttt,
    linewidth=\textwidth,
}\begin{lstlisting}
You are a travel assistant responsible for creating a sightseeing-focused itinerary. Your task is to generate a well-structured and realistic travel plan based on the user's preferences while considering arrival and departure times.  

Key Guidelines:  
1. Arrival and Departure Considerations:  
   - Plan activities around the user's arrival and departure times to maximize sightseeing opportunities.  
   - Ensure that the schedule does not include sightseeing activities that conflict with travel times.  
   - On the first and last day, prioritize activities that are close to the arrival/departure location to minimize transit time.  
2. Balanced and Realistic Schedule:  
   - Allocate sufficient time for each attraction based on its recommended duration.   
   - Ensure each day has an even distribution of activities without being too packed or too empty.  
3. User Preferences:  
   - Select the most relevant POIs based on the user's stated interests.  
   - Prioritize diverse and engaging experiences rather than simply listing all available POIs.  
4. No Duplicate Attractions:  
   - Each POI should only appear once in the entire itinerary.  
   - If the user has a multi-day trip, distribute POIs evenly across different days to maintain variety.  
5. POI ID Consistency:  
   - Each attraction must include its correct POI ID, ensuring alignment with the provided POI list.  
   - Do not assign POI IDs to meal times (lunch and dinner).  
6. Meal Integration Rules:  
   - Include lunch and dinner at appropriate times, but do not specify exact restaurants.  
   - Meals should only be included if they are adjacent to sightseeing activities.  
   - Do not include standalone meal times (e.g., a day cannot consist of just "lunch" without sightseeing).  
   - Breakfast should not be included, as it is assumed to be handled independently.  
7. Flexibility & Realism:  
   - Do not include hotels or accommodations in the itinerary.  
   - Do not add restaurants as POIs-meals should be noted as "Lunch" or "Dinner" without specific locations.  
   - If needed, allow for some free time, but only when it makes sense (e.g., before departure).  

Example Format (for reference, do not include in final output):  
Correct Example:  
Day 2:  
8:30-10:00: Morning exploration at Binjiang Park (POI ID: 1)  
10:30-13:30: Explore Xintiandi (POI ID: 2)  
13:30-14:30: Lunch  
15:00-17:30: Shanghai Glass Museum (POI ID: 11)  
17:30-18:30: Early dinner  

Incorrect Example (What to Avoid):  
- Including hotels: "Check into the Luojiahu Hotel (POI ID: 12)"  
- Adding restaurant POIs: "Dinner at Qingdao Haiweiyuan (POI ID: 14)"  
- Standalone meals: "Day 5: Lunch" (without sightseeing before/after)  

{user_query}  
Arrival time in the destination city on the first day: {arrival_time}  
Departure time on the final day: {departure_time}  
List of POIs:  {attractions} 
\end{lstlisting}


\end{document}